\documentclass[11pt,a4paper]{article}
\usepackage[hyperref]{emnlp2020}
\usepackage{times}
\usepackage{latexsym}

\aclfinalcopy

\usepackage{latexsym}
\usepackage{soul}
\usepackage{amsfonts}
\usepackage{amsmath}
\usepackage{amssymb}
\usepackage{mathptmx}
\usepackage{nicefrac}
\usepackage{booktabs} % Pretty tables
\usepackage{makecell} % cell line breaks..
\usepackage{graphicx}
\usepackage[inline]{enumitem}
\usepackage[font=small]{subcaption}
\usepackage[font=small]{caption}
\usepackage{xargs} % commandx
\usepackage{color}
\usepackage[textsize=scriptsize]{todonotes}
\definecolor{blue}{RGB}{0, 93, 170}			%Go Big Blue!
\definecolor{darkgreen}{RGB}{0, 102, 0}

\newcommand{\ignore}[1]{}
\newcommand*\samethanks[1][\value{footnote}]{\footnotemark[#1]}
\title{Enhancing Text-based Reinforcement Learning Agents\\ with Commonsense Knowledge}

\author{Keerthiram Murugesan \\
\texttt{keerthiram.murugesan} \\ \texttt{@ibm.com}\\\And 
Mattia Atzeni\thanks{Both student authors contributed equally.} \\
\texttt{atz@zurich.ibm.com}\\\And 
Pushkar Shukla\samethanks \\
\texttt{pushkarshukla@ttic.edu}\\\AND
Mrinmaya Sachan \\
\texttt{mrinmaya@ttic.edu}\\\And
Pavan Kapanipathi \\
\texttt{kapanipa@us.ibm.com}\\\And
Kartik Talamadupula\\
\texttt{krtalamad@us.ibm.com} }
\date{}

\begin{document}

\maketitle

\begin{abstract}

In this paper, we consider the recent trend of evaluating progress on reinforcement learning technology by using text-based environments and games as evaluation environments. This reliance on text brings advances in natural language processing into the ambit of these agents, with a recurring thread being the use of external knowledge to mimic and better human-level performance. We present one such instantiation of agents that use commonsense knowledge from ConceptNet to show promising performance on two text-based environments.

\end{abstract}

\section{Introduction}
\label{sec:intro}
Over the years, simulation environments and games have been used extensively to showcase and drive advances in reinforcement learning technology. A recent environment that has received much focus is TextWorld (TW)~\cite{cote18textworld}, where an agent must interact with an external environment to achieve goals while maximizing reward - all of this using only the modality of text. TextWorld and similar text-based tasks seek to bring advances in natural language processing (NLP) and question answering solutions to agent-based reinforcement learning techniques, and vice-versa.

\begin{figure*}[tp]
    \centering
    \includegraphics[width=0.8\linewidth]{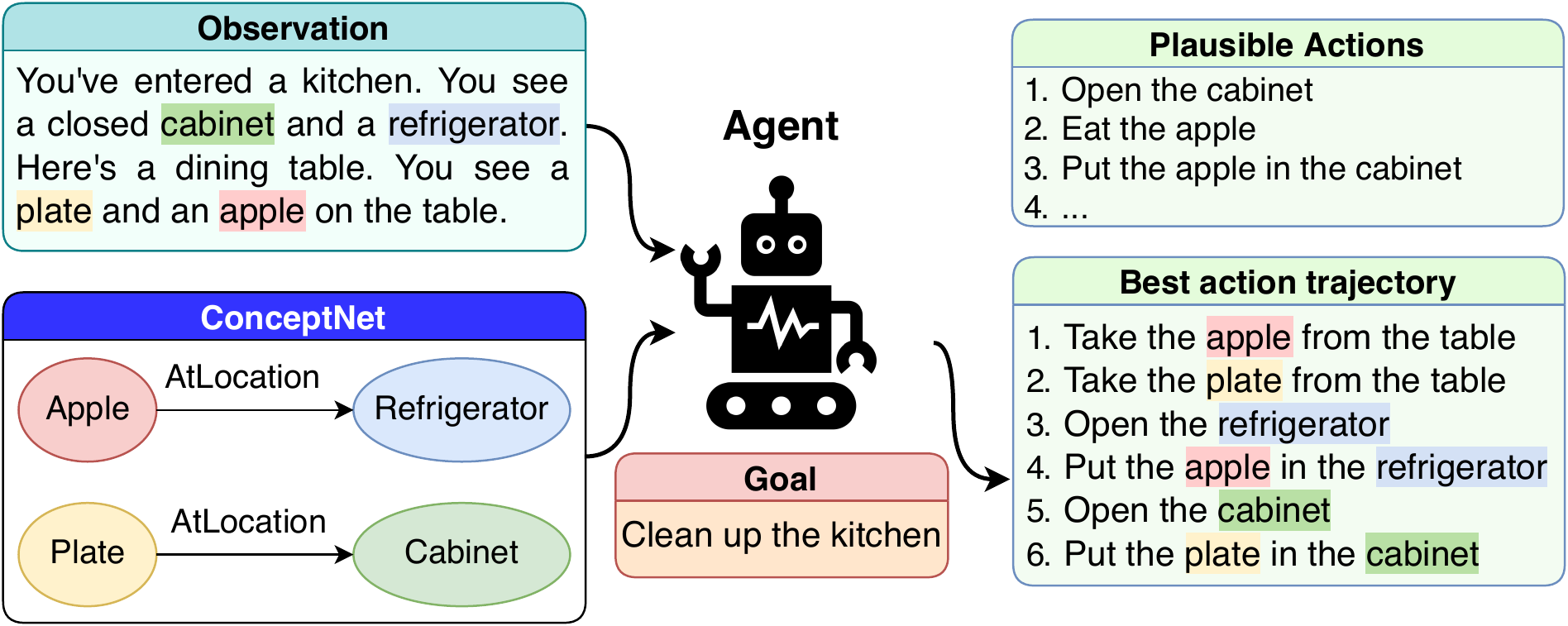}
    \caption{An illustration of our Kitchen Cleanup game. The agent perceives the world via text and has been given the goal of cleaning up the kitchen. As shown here, the agent can leverage commonsense knowledge from ConceptNet to reduce the exploration and achieve the goal.}
    \label{fig:kitchen_cleanup}
\end{figure*}

% \mrinmaya{For the below paragraph, I would suggest talking about how commonsense knowledge can make RL more efficient citing Figure 1. KRT: Resolved}
A common thread inherent in solutions to some of the NLP tasks is that mere text-based techniques cannot achieve or beat the human-level performance and that NLP systems must instead learn how to utilize additional knowledge from external sources such as knowledge bases (KBs) and knowledge graphs (KGs) to improve their overall performance. Figure~\ref{fig:kitchen_cleanup} presents a running example that illustrates this: in the figure, the additional knowledge that must be utilized effectively by the agent is presented in the bottom left corner under the {\em ConceptNet} heading.

In general, the use of external knowledge to improve the accuracy of NLP tasks has garnered significant attention from the community. Specifically, for tasks like natural language inference (NLI), recent work~\cite{kapanipathi2020infusing,wang2019improving} has shown that while external knowledge can bring in useful information, this must be balanced by the context-specific relevance of the new information fed into the system. If this is not done properly, there is a very high risk of overwhelming the agent/algorithm with too much information, leading to poor decisions and performance.

In this paper, we present a novel approach to the use of external knowledge from the ConceptNet~\cite{liu2004conceptnet,speer2017conceptnet} knowledge graph to reduce the exploration space for a Reinforcement Learning (RL) agent. Specifically, we consider an RL based agent that is able to model the world around it at two levels -- a {\em local}, or {\em belief}, graph that describes its current belief of the state of the world; and a {\em global} or {\em commonsense} graph of entities that are related to that state -- and the interaction between those two levels. 
The belief graph provides the agent with a symbolic way to represent its current perception of the world, which can be easily combined with symbolic commonsense knowledge from the commonsense graph. This two-level representation of the world and the knowledge about it follows the model proposed in the Graph Aided Transformer Agent (GATA)~\cite{adhikari2020learning} framework. 

Using this model, we are able to show a significant improvement in the performance of an RL agent in a kitchen cleanup task that is set in the TextWorld setting. An example of such a kitchen cleanup task is shown in Figure~\ref{fig:kitchen_cleanup}: the agent is given an initial observation (which is used to produce the first iteration of the agent's belief graph), with the final goal of cleaning up the kitchen. The agent has to produce the list of actions that are necessary to achieve this goal: that list is given on the right hand side. Finally, the additional external knowledge from the ConceptNet knowledge graph -- which makes up the global graph for our agent -- is shown at the bottom left. In the case of this running example, the agent may discover from ConceptNet that apples are usually located in refrigerators, and plates are located in cabinets. We will use this kitchen cleanup instance as a running example throughout the paper.

By evaluating our approach on two different tasks -- a {\em kitchen cleanup} task as above, and an additional {\em cooking recipe} task -- we can show that the interaction between the belief and commonsense graphs can reduce the exploration of the RL agent in comparison to the purely text-based model. However, we are also able to demonstrate a more nuanced point: merely providing an agent with commonsense knowledge is not sufficient to improve its performance. Indeed, oftentimes it is detrimental to the agent's performance. We show that this is due to the agent being overwhelmed with too much commonsense knowledge, and discuss how different tasks and settings have different demands on the knowledge that is used by an agent.

% % In this paper, we present the first true use of knowledge-based embeddings from open knowledge sources like ConceptNet in order to improve the performance of a reinforcement learning agent.

% \begin{itemize}
% \item Novel approach to use external knowledge to reduce exploration for an RL Agent. We demostrate the impact of it on Text-World setting.
% \item First one to look at the interaction between Local and Global graphs (GATA paper) and their interaction
% \item We show that the use of external knowledge, in particular the interaction between local and global graphs, reduces the exploration of the RL agent by X\% in comparison to the pure text based.
% \end{itemize}
\section{Related Work}

We start out with a look at work that is related to our focus area, which we categorize into three primary areas below. Our work sits at the intersection of knowledge graphs and the use of commonsense (and external) knowledge to make reinforcement learning more efficient; and our improvements are showcased in TextWorld and adjacent text-based domains.

% We will categorize the related work into three primary areas

% \begin{itemize}
% \item IQA: Textworld and textworld-adjacent work (Kartik)
% \item Reinforcement Learning as applied to IQA-like tasks (slow, convergence, etc.) (RL for IQA -- Kartik and RL for KG -- Pavan )
% \end{itemize}

\subsection{Knowledge Graphs}
\label{subsec:related_knowledge_graphs}

% \mrinmaya{I feel 2.1 is too long; KRT: Agreed, I can take a pass at trimming it down TODO}

Graphs have become a common way to represent knowledge. 
These {\em knowledge graphs} consist of a set of concepts (nodes) connected by relationships (edges). Well-known knowledge graphs (KGs) that are openly available include Freebase~\cite{freebase}, DBpedia~\cite{auer2007dbpedia}, WordNet~\cite{wordnet}, and ConceptNet~\cite{speer2017conceptnet}. Each of these KGs comprises different types of knowledge.
% (1) DBPedia and Freebase typically contain encyclopedic knowledge either crowd sourced or derived from Wikipedia; 
% % For example: \texttt{dbr:Barack\_Obama} $\rightarrow$ \texttt{dbo:spouse}  $\rightarrow$ \texttt{dbr:Michelle\_Obama} 
% (2) WordNet has linguistic knowledge such as synonyms, antonyms, and hypernyms; 
% % (\texttt{sprint} $\rightarrow$ \texttt{synonym} $\rightarrow$ \texttt{run}).   
% (3)  ConceptNet is a commonsense knowledge base containing information such as \texttt{Refrigerator}  $\rightarrow$ \texttt{AtLocation}  $\rightarrow$ \texttt{Kitchen}. 
For the tasks that our work considers, we found that the commonsense knowledge available in ConceptNet is more suitable than the encyclopedic knowledge from DBpedia or Freebase -- we hence focus on this. Since our approach considers the KG as a generic graph structure, it is amenable to the use of any of the KGs mentioned here.

% However, our approach considers the knowledge graph as a graph structure, hence is flexible to leverage any knowledge graphs mentioned above. 

Knowledge graphs have been used to perform reasoning to improve performance in various domains, particularly within the NLP community. In particular, KGs have been leveraged for tasks such as Entity Linking~\cite{hoffart2012kore}, Question Answering~\cite{sun2018open,das2017go,atzeni2018what}, Sentiment Analysis~\cite{reforgiato2015,atzeni2018using} and Natural Language Inference~\cite{kapanipathi2020infusing}. Different techniques have been explored for their use. In most cases, knowledge graph embeddings such as TransH~\cite{transh} and ComplEx~\cite{complex} are used to vectorize the concepts and relationships in a KG as input to a learning framework. Reinforcement learning has also been used to find relevant paths in a knowledge graph for knowledge base question answering~\cite{das2017go}. \citet{sun2018open} and \citet{kapanipathi2020infusing} find sub-graphs from the corresponding KGs and encode them using a graph convolutional networks~\cite{kipf2016semi} for question answering and natural language inference respectively.

%Such facts in knowledge graphs can be leveraged to augment text by expanding the concepts in the text to include related concepts from the knowledge graph. Techniques that augment text with concepts from knowledge graphs have already been shown to be beneficial~\cite{hoffart2012kore,kapanipathi2014user}\pavan{(add relevant citations)}. Inspired by such works, this paper explores harnessing knowledge graphs for NLI. 

\subsection{External Knowledge for Sample Efficient Reinforcement Learning}
\label{subsec:ext_knowledge_sample_efficient_rl}

% \mrinmaya{I started this section to have a more concrete discussion of previous work in RL+commonsense - to improve sample efficiency? Maybe we can borrow some of this discussion in the intro to carve out our contribution.}

A key challenge for current reinforcement learning (RL) technology is the low sample efficiency \cite{kaelbling1998planning}. RL techniques require a large amount of interaction with the environment which can be very expensive. This has prevented the use of RL in real-world decision-making problems. In contrast, humans possess a wealth of \textit{commonsense} knowledge which helps them solve problems in the face of incomplete information.

Inspired by this, there have been a few recent attempts on adding prior or external knowledge to RL approaches. Notably, \citet{garnelo2016towards} propose \textit{Deep Symbolic RL}, which combines aspects of symbolic AI with neural networks and reinforcement learning as a way to introduce common sense priors. However, their work is mainly theoretical. There has also been some work on \textit{policy transfer}~\cite{bianchi2015transferring}, which studies how 
knowledge acquired in one environment can be re-used in another environment;
and \textit{experience replay}~\cite{wang2016sample,lin1992self,lin1993reinforcement} which studies how an agent's previous experiences can be stored and then later reused.
In contrast to the above, in this paper, we explore the use of commonsense knowledge stored in knowledge graphs such as ConceptNet as a way to improve sample efficiency in text-based RL agents. To the best of our knowledge, there is no prior work that explores how commonsense knowledge can be used to make RL agents more efficient.

%Papers to cite and contrast \cite{juba2016integrated} \cite{garcez2018towards}

\subsection{RL Environments and TextWorld}
\label{subsec:rl_environments_tw}

Games are a rich domain for studying grounded language and how information from the text can be utilized in controlled applications. Notably, in this line of research, \citet{branavan2012learning} builds an RL-based game player that utilizes text manuals to learn strategies
for Civilization II; and \citet{narasimhan2015language} build an RL-based game player for multi-user Dungeon games. In both cases, the text is analyzed and control strategies are learned jointly using feedback from the gaming environment. Similarly, in the vision domain, there has been work on building automatic video game players~\cite{koutnik2013evolving,mnih2016asynchronous}.

Our work builds on a recently introduced text-based game \textit{TextWorld}~\cite{cote18textworld}.
TextWorld is a sandbox learning environment for training and evaluating RL-based agents on text-based games. Since its introduction and other such tools, there has been a large body of work devoted to improving performance on this benchmark.
One interesting line of work on TextWorld is on learning symbolic (typically graphical) representations of the agent's belief of the state of the world. Notably, \citet{ammanabrolu2019playing} proposed \textit{KG-DQN} and \citet{adhikari2020learning} proposed \textit{GATA}; both represent the game state as a belief graph learned during exploration. This graph is used to prune the action space, enabling more efficient exploration. Similar approaches for building dynamic belief graphs have also been explored in the context of machine comprehension of procedural text~\cite{Das:2018}.
In our work, we also represent the world as a belief graph. Moreover, we also explore how the belief graph can be used with commonsense knowledge for efficient exploration.

The \textit{LeDeepChef} system~\cite{Adolphs2019LeDeepChefDR}, which investigates the generalization capabilities of text-based RL agents as they learn to transfer their cooking skills to never-before-seen recipes in unfamiliar house environments, is also related to our work. They achieve transfer by additionally supervising the model with a list of the most common food items in Freebase, allowing their agent to generalize to hitherto unseen recipes and ingredients.

%\begin{itemize}
    %\item GATA \cite{adhikari2020learning} generates dynamic graph representation from the raw text observation to facilitate planning and generalization. Uses  a novel transformer-based sequence-to-sequence model that constructs a “belief” KG and updates this graph dynamically
    %\item Le Deep Chef \cite{Adolphs2019LeDeepChefDR} outperformed in FirstTextWorld competition with several but similar cooking-related tasks in a modern house environment
    %\item 

Finally, \citet{zahavy2018learn} propose the Action-Elimination Deep
Q-Network (AE-DQN) which learns to predict invalid actions in the text-adventure game \textit{Zork}, and eliminates them using contextual bandits. This allows the model to efficiently handle the large action space. The use of common sense knowledge in our work potentially has the same effect of down-weighting implausible actions.

\section{TextWorld as a POMDP}
\label{sec:textworld_as_pomdp}

Text-based games can be seen as partially observable Markov decision processes (POMDP) \cite{kaelbling1998planning} where the system dynamics are determined by an MDP, but the agent cannot directly observe the underlying state.
As an agent interacts with a TextWorld game instance, at each turn, several lines of text describe the state of the game; and the player can issue a text command to change the state in some desirable way (typically, in order to move towards a goal).

Formally, let $(S, T, A,\Omega, O, R, \gamma)$ denote the underlying TextWorld POMDP. Here, $S$ denotes the set of states, $A$ denotes the action space, $T$ denotes the state transition probabilities, $\Omega$ denotes the set of observations, $O$ denotes the set of conditional observation probabilities, and $\gamma \in [0, 1]$ is the discount factor.
The agent's observation $o_t$ at time step $t$ depends on the current state $s_t$ and the previous action $a_{t-1}$.
The agent receives a reward at time step $t$: $r_t=R(s_t,a_t)$ and the agent's goal is to maximize the expected discounted sum of rewards: 

\[\mathbb{E}[\sum_t\gamma^{t}\ r_t].\]

\noindent TextWorld allows the agent to perceive and interact with the environment via the modality of text. Thus, the observation $o_t$ is presented by the environment as a sequence of tokens ($o_t =  \{o_t^1, \dots o_t^N\}$). Similarly, each action $a$ is also denoted as a sequence of tokens $\{a^1, \dots, a^M\}$.

\section{Model Description}
\begin{figure*}
    \centering
    \includegraphics[width=\linewidth]{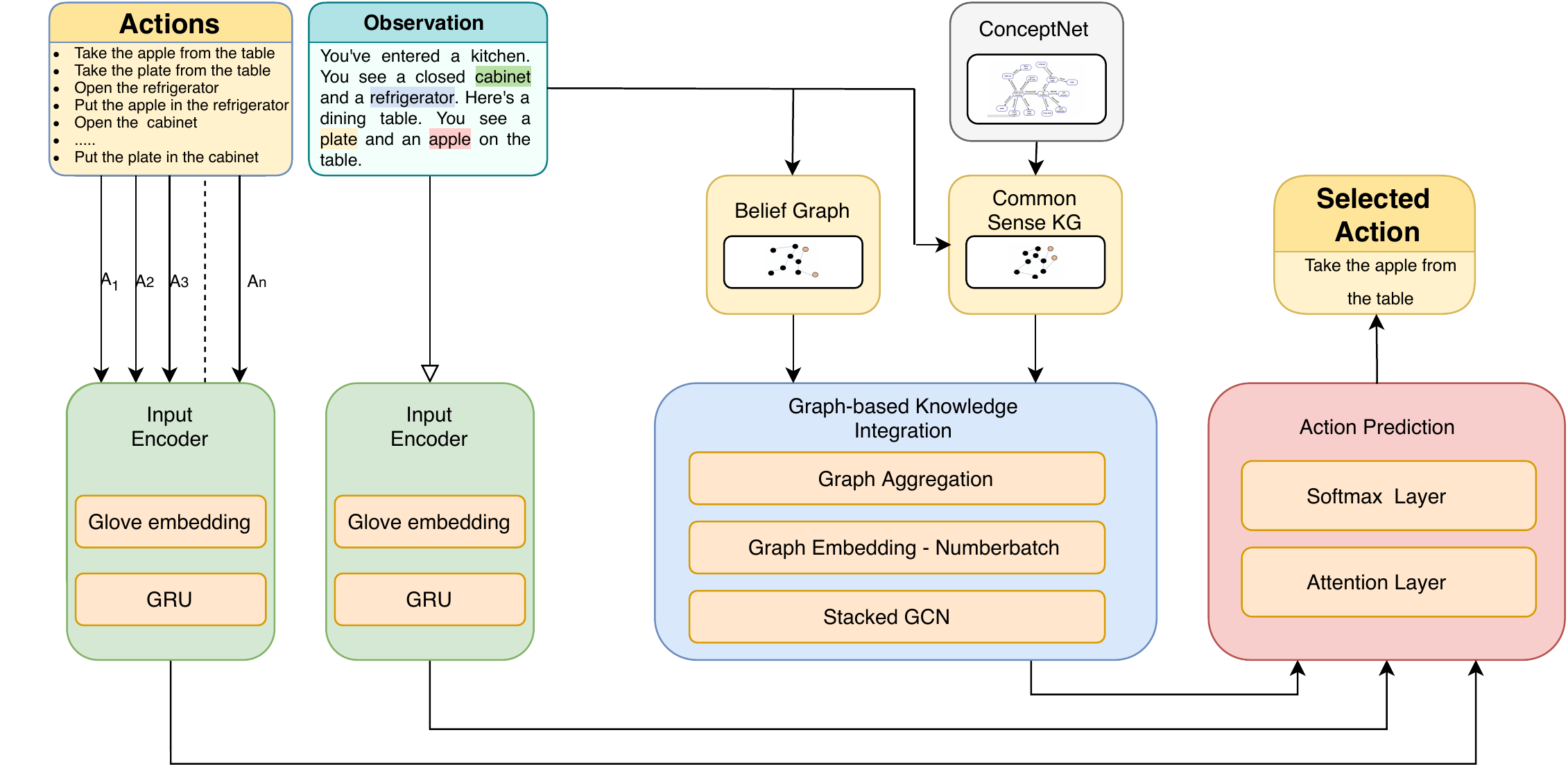}
    \caption{Overview of our model's decision making at any given time step. Our model comprises of the following components: (a) an input encoder, which encodes admissible actions and observations, (b) belief graph, which captures the agent's belief of the current state, (c) common sense KG, a subgraph of ConceptNet extracted by the agent, (d) knowledge integration of the belief graph and the extracted common sense KG, and (e) action prediction.}
    \label{fig:block_diagram}
\end{figure*}

In order to solve the above POMDP, we design a model that can leverage commonsense knowledge and learn a graph-structured representation of its belief of the world state. The high-level architecture of the model contains three major components, namely the input encoder, a graph-based knowledge extractor, and the action prediction module. The input encoding layers are used for encoding the observation at time step $t$ and the list of admissible actions. The graph-based knowledge extractor tries to extract knowledge from two different sources. First, it makes use of external commonsense knowledge to improve the ability of the agent to select the correct action at each time step. Secondly, the belief about the environment (world state) perceived by the agent is also captured by a belief graph that is generated dynamically from the textual observations from the game. The information from both sources is then aggregated together in a single graph. The action-prediction module takes as input the encoded observation states, the encoded list of admissible actions and the encoded aggregated graph, and predicts an action for each step. Figure \ref{fig:block_diagram} provides a compact visualization of our approach. We describe the various components of our model below.

\subsection{Input Encoder}
\label{sec:input_encoder}
% The input embedding layer is used for encoding the observation and list of available actions. \\
At any time step $t$, the agent observes a textual description of the current state provided as a sequence of tokens $o_t = (o_t^1, \dots, o_t^N)$.
% At any given instance,the observation $o_t$ is represented with $d$-dimensional glove embeddings $(\mathbf{x}_t^1, \dots, \mathbf{x}_t^N)$ where $\mathbf{x}_t^k \in \mathbb{R}^d$ is the glove embedding of token $o_t^k$. Further, the glove embedding for the observation is passed through a GRU encoder such that
% $\mathbf{o}_t = \mathbf{h}^N_{t}$, where $\mathbf{h}^k_{t} = GRU(\mathbf{h}^{k - 1}_{t}, \mathbf{x}_t^k)$
% We use a similar procedure to encode the list of available actions. Each action $a_i \in A_t$ is represented as $(\mathbf{c}_i^1, \dots, \mathbf{c}_i^N)$ where $\mathbf{c}_i^k  \in \mathbb{R}^d$ is the glove embedding of token $a_i^k$. The embedding is then passed through a GRU encoder such that
% $\mathbf{a}_i = \mathbf{a}_i^M$, where $\mathbf{a}_i^k = GRU(\mathbf{a}_i^{k-1}, \mathbf{c}_i^k)$
Given the current observation $o_t$, we use pre-trained GloVe embeddings \cite{pennington2014glove} to represent $o_t$ as a sequence of $d$-dimensional vectors $\mathbf{x}_t^1, \dots, \mathbf{x}_t^N$, where each $\mathbf{x}_t^k \in \mathbb{R}^d$ is the glove embedding of the $k$-th observed token $o_t^k$, $k = 1, \dots, N$.
Similarly, given the set $A_t$ of admissible actions at time step $t$, we represent each action $a_{i} = (a_{i}^1, \dots, a_{i}^M) \in A_t$ as a sequence of $d$-dimensional pretrained GloVe embeddings $\mathbf{c}_i^1, \dots, \mathbf{c}_i^M$. 

The agent relies on a hierarchical encoder architecture to model the current state as a vector $\mathbf{s}_t$, based on $o_t$ and the previous observations. First, a GRU-based encoder is used to process the sequence $\mathbf{x}_t^1, \dots, \mathbf{x}_t^N$ of the GloVe embeddings associated with $o_t$. This allows representing the current observation as a single $h$-dimensional vector $\mathbf{o}_t \in \mathbb{R}^h$, where $h$ is the output dimensionality of the GRU. Formally, $\mathbf{o}_t$ is computed as  $\mathbf{o}_t = \mathbf{h}^N_{t}$, with $\mathbf{h}^k_{t} = GRU(\mathbf{h}^{k - 1}_{t}, \mathbf{x}_t^k)$, for $k = 1, \dots, N$. In the previous equation, $GRU(\cdot)$ refers to the forward propagation of a gated recurrent unit \cite{cho2014gru}.
Then, the sequence of previous observations up to $o_t$ is encoded in a similar way into a vector $\mathbf{s}_t = GRU(\mathbf{s}_{t-1}, \mathbf{o}_t) \in \mathbb{R}^h$.
We do the same to represent each admissible action $a_i$ as $\mathbf{a}_i = \mathbf{a}_i^M$, with $\mathbf{a}_i^k = GRU(\mathbf{a}_i^{k-1}, \mathbf{c}_i^k)$, for $k = 1, \dots, M$.

\subsection{Graph-based Knowledge Integration}
We enhance our text-based RL agent by allowing it to access a graph that captures both commonsense knowledge and the agent's current belief of the world state.
Formally, we assume that, at each time step $t$, the agent has access to a graph $G_t = (\mathcal{V}_t, \mathcal{E}_t)$, where $\mathcal{V}_t$ is the set of nodes and $\mathcal{E}_t \subseteq \mathcal{V}_t^2$ denotes the edges of the graph.
The graph is updated dynamically at each time step $t$ and new nodes are either added or deleted based on the textual observation $o_t$.

As mentioned, $G_t$ encodes both commonsense knowledge and the belief of the world state. Commonsense knowledge is extracted from the history of the observations by linking the entities mentioned in the text to an external KG. This allows extracting a \textit{commonsense knowledge graph}, which is a subgraph of the external source of knowledge providing information about the entities of interest. In our experiments, we use ConceptNet \cite{speer2017conceptnet} as the external knowledge graph.
On the other hand, the observation $o_t$ is also used to update a dynamically generated \textit{belief graph} as in recent work by \citet{adhikari2020learning}. The graph aggregation is performed by merging the belief and commonsense knowledge graphs based on the entity mentions. This helps to reduce the noise extracted from updating both the belief and the commonsense graphs. 
As shown in Figure \ref{fig:block_diagram}, the commonsense knowledge graph, and the belief graph are updated based on the observations, and then they are aggregated to form a single graph $G_t$. 
The graph $G_t$ at time step $t$ is processed by a graph encoder as follows. First, pretrained KG embeddings are used to map the set of nodes $\mathcal{V}_t$ into a feature matrix 

\[ E_t = [\mathbf{e}_t^1, \dots, \mathbf{e}_t^{|\mathcal{V}_t|}] \in \mathbb{R}^{f \times |\mathcal{V}_t|},\] 

\noindent where each column $\mathbf{e}_t^i \in \mathbb{R}^f$ is the embedding of node $i \in \mathcal{V}_t$. We use Numberbatch embeddings \cite{speer2017conceptnet} to create the matrix $E_t$. Such feature matrix provides initial node embeddings that are iteratively updated by message passing between the nodes of $G_t$, using $L$ stacked GCN (graph convolutional network) layers \cite{kipf2016semi}, where $L$ is an hyperparameter of the model. The output of this process is an updated matrix $Z_t = [\mathbf{z}_t^1, \dots, \mathbf{z_t^{|\mathcal{V}|}}] \in \mathbb{R}^{h \times |\mathcal{V}_t|}$.
We then compute a graph encoding $\mathbf{g}_t$ for $G_t$ by simply averaging over the columns of $Z_t$, namely:

\[ \mathbf{g}_t = \frac{1}{|\mathcal{V}_t|} \sum_{i = 1}^{|\mathcal{V}_t|}\mathbf{z}_t^i. \]

In our experiments, we use the updated KG embeddings to create a graph-based encoding vector for each action as described in Section \ref{sec:input_encoder}, in addition to the graph encoding $\mathbf{g}_t$. This approach has shown to be a better integration of the knowledge graph at each time step.  

\subsection{Action Prediction}
%The original observation and action embeddings ,the embeddings for the list of available actions, the output of the graph aggregator are all concatenated together and act as an input to the Action Prediction Module. The layer is followed by an attention module. The concatenated representations are fed into a 2-layer MLP with a RELU non-linear activation function in between. The output of this layer gives us the score value for each command.

The action $\hat{a}_t$ selected by the agent at time step $t$ is computed based on the state embedding $\mathbf{s}_t$, the graph embedding $\mathbf{g}_t$ and the action candidates $\mathbf{a}_1, \dots, \mathbf{a}_{|A_t|}$.
First, all these vectors are concatenated together into a single vector $\mathbf{r}_t = [\mathbf{g}_t; \mathbf{s}_t; \mathbf{a}_1; \dots; \mathbf{a}_{|A_t|}]$.
Then, we compute a vector $\mathbf{p}_t \in \mathbb{R}^{|A_t|}$ with a probability score for each action $a_i \in A_t$ as:

\[ \mathbf{p}_t = softmax(W_1 \cdot ReLU(W_2 \cdot \mathbf{r}_t + \mathbf{b}_2) + \mathbf{b}_1) \] 

\noindent where $W_1, W_2, \mathbf{b}_1$, and $\mathbf{b}_2$ are learnable parameters of the model. The final action chosen by the agent is then given by the one with the maximum probability score, namely $\hat{a}_t = \arg \max_i p_{t,i}$.

\subsection{Learning}
Following the winning strategy in the FirstTextWorld competition \cite{Adolphs2019LeDeepChefDR}, we use the Advantage Actor-Critic (A2C) framework \cite{mnih2016asynchronous} to train the agent and optimize the action selector on reward signals from training games.
\section{Experiments}
\label{sec:experiments}

\begin{figure*}
    \centering
        \includegraphics[scale=0.6]{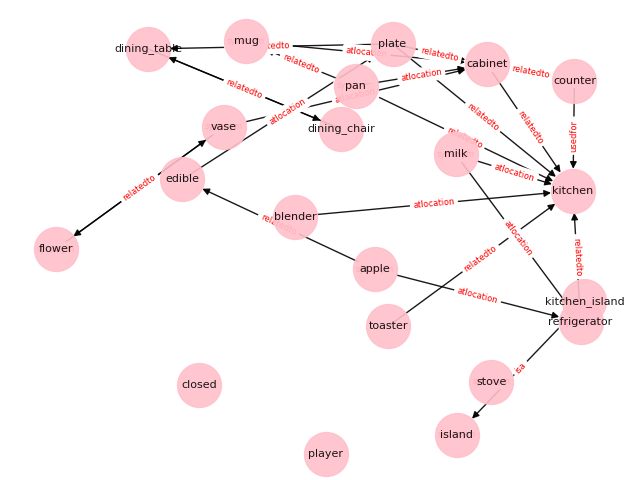}
    \caption{Example of the commonsense knowledge graph extracted from ConceptNet for the Kitchen Cleanup Task}
    \label{fig:kitchen_graph}
\end{figure*}

In this section, we report on experiments to study the role of commonsense knowledge-based RL agents in the TextWorld environment. We evaluate and compare our agent on two sets of game instances: 1) Kitchen Cleanup Task, and 2) Cooking Recipe Task.

\subsection{Kitchen Cleanup Task}

First, we use  TextWorld \cite{cote2018textworld} to generate a game/task to assess the performance gain using commonsense knowledge graphs such as ConceptNet. We generate the game with $10$ objects relevant to the game, and $5$ distractor objects spread across the room. The goal of the agent is to tidy the room (kitchen) by putting the objects in the right place. We create a set of realistic kitchen cleanup goals for the agent: for instance, \textit{take apple from the table} and \textit{put apple inside the refrigerator}.  Since information on concepts that map to the objects in the room is explicitly provided in ConceptNet (\texttt{Apple}  $\rightarrow$ \texttt{AtLocation}  $\rightarrow$ \texttt{Refrigerator}), the main hypothesis underlying the creation of this game is that leveraging the commonsense knowledge could allow the agent to achieve a higher reward while reducing the number of interactions with the environment.

The agent is presented with the textual description of a kitchen, consisting of the location of different objects in the kitchen and their spatial relationship to the other objects. The agent uses this information to select the next action to perform in the environment. Whenever the agent takes an object and puts it in the target location, it receives a reward and its total score goes up by one point. The maximum score that can be achieved by the agent in this kitchen cleanup task is equal to $10$.  In addition to the textual description, we extract the commonsense knowledge graph from ConceptNet based on the text description.  Figure~\ref{fig:kitchen_graph} shows an instance of the commonsense knowledge graph created during the agent's interaction with the environment. Note that even for the simple kitchen cleanup task that we model (see Figure~\ref{fig:kitchen_cleanup} for details), the commonsense knowledge graph contains more than $20$ entities (nodes) and a similar number of relations (edges). This visualization is useful, as it lends a basis for our upcoming discussion on agents being overwhelmed with {\em too much} commonsense knowledge.

\begin{figure*}[t]
    \centering
    \includegraphics[scale=0.48]{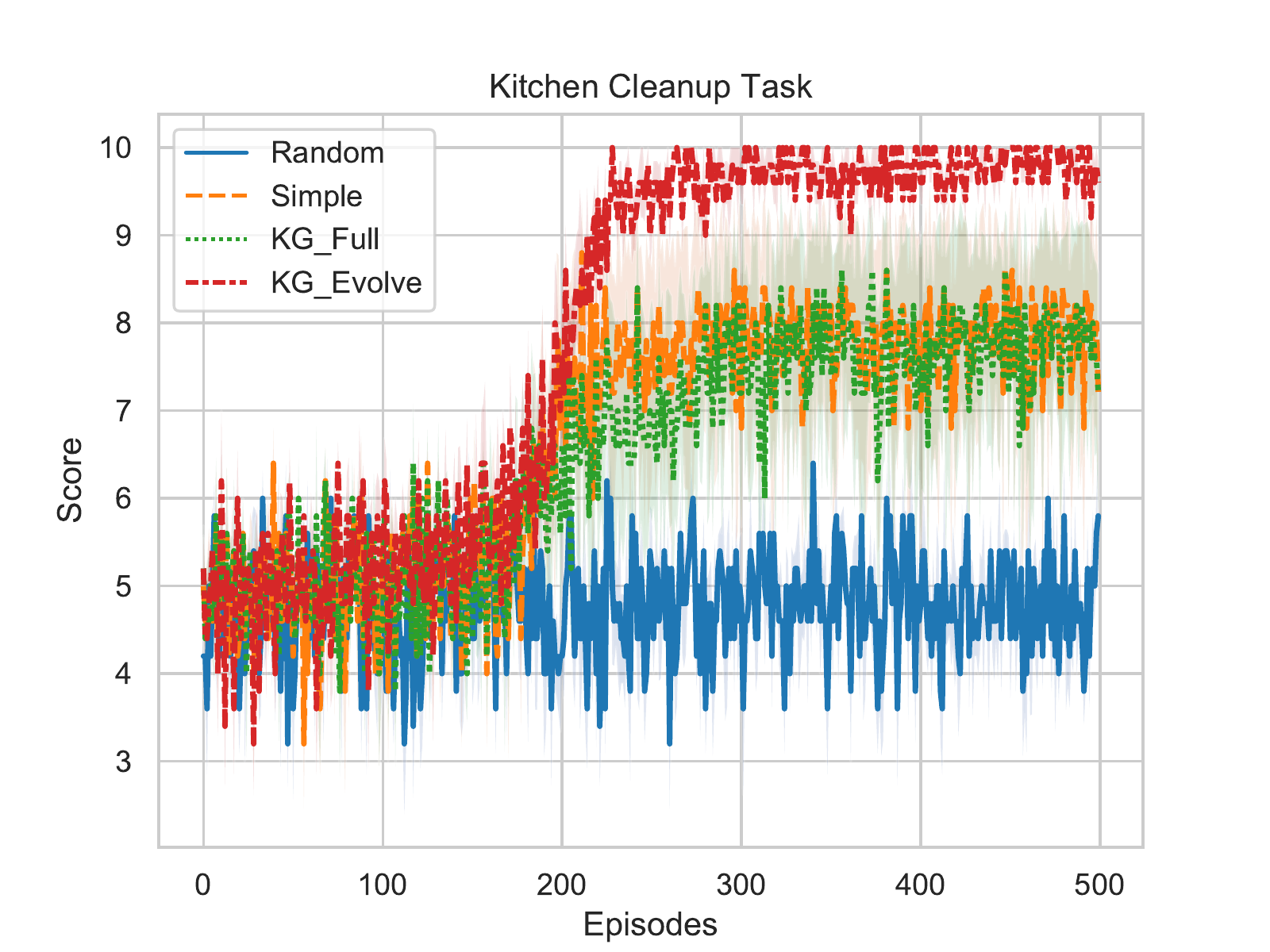}
    \includegraphics[scale=0.48]{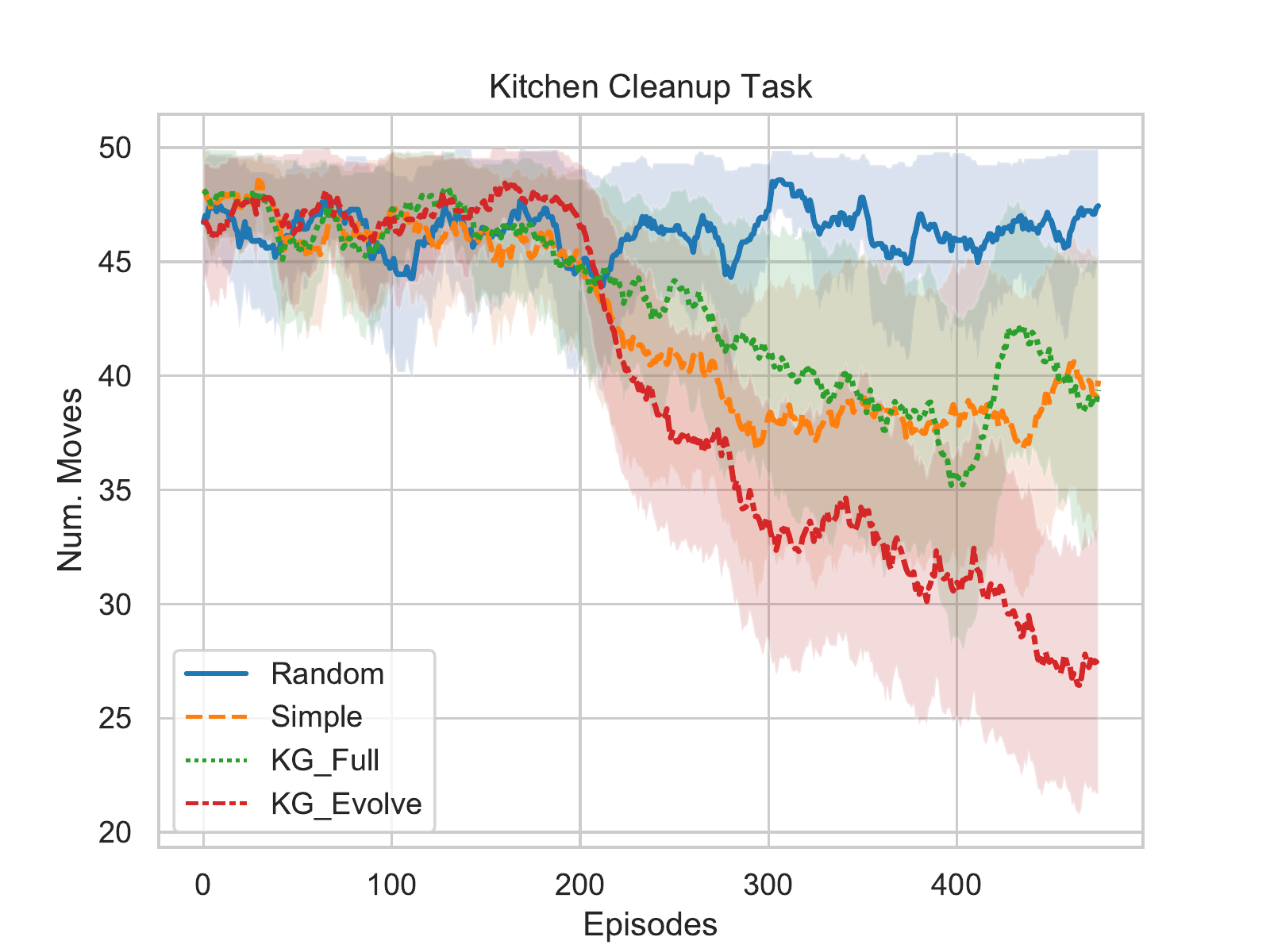}
    \caption{Comparison of agents for the Kitchen Cleanup task with and without commonsense knowledge (Conceptnet) with average scores and average moves (averaged over 5 runs).}
    \label{fig:kitchen_scores}
\end{figure*}

\subsubsection{Results on Kitchen Cleanup}

We compare our knowledge-aware RL agents (\textit{KG\_Full} and \textit{KG\_Evolve}) against two baselines for performance comparison: \textit{Random}, where the agent chooses an action randomly at each step; and \textit{Simple}, where the agent chooses the next action using the text description alone and ignores the commonsense knowledge graph. The knowledge-aware RL agents, on the other hand, use the commonsense knowledge graph to choose the next action. The graph is provided in either full-graph setting where all the commonsense relationships between the objects are given at the beginning of the game (\textit{KG\_Full}); or evolve-graph setting where only the commonsense relationship between the objects seen/interacted by the agent until the current steps are revealed (\textit{KG\_Evolve}).  We record the average score achieved by each agent and the average number of interactions (moves) with the environment as our evaluation metrics. Figure~\ref{fig:kitchen_scores} shows the results for the kitchen cleanup task averaged over $5$ runs, with $500$ episodes per run.

\subsubsection{Discussion of Kitchen Cleanup}

As expected, we see that agents that use the textual description and additionally the commonsense knowledge outperform the baseline random agent. We are also able to demonstrate clearly that the knowledge-aware agent outperforms the simple agent with the help of commonsense knowledge.  The knowledge-aware agent with the evolve-graph setting outperforms both the simple agent as well as the agent with the full-graph setting. We believe that when an agent has access to the full commonsense knowledge graph at the beginning of the game, the agent gets overwhelmed by the amount of knowledge given; and is prone to making noisy explorations in the environment. On the other hand, feeding the commonsense knowledge gradually during the agent's learning process provides more focus to the exploration, and drives it toward the concepts related to the rest of the goals. These results can also be seen as an RL-centric agent-based validation of similar results shown in the broader NLP literature by the work of \cite{kapanipathi2020infusing}.

\subsection{Cooking Recipe Task}
\label{subsec:cooking_recipe_experiments}

Next, we evaluate the performance of our agent on the cooking recipe task by using $20$ different games generated by \cite{adhikari2020learning}. These games follow a recipe-based cooking theme, with a single ingredient in a single room (difficulty level 1). The goal is to collect that specific ingredient to prepare a meal from a given recipe. 

As in our previous task, we compare our agent with the \textit{Simple} agent. In addition to the simple agent, we compare our agent with the \textit{GATA} agent \cite{adhikari2020learning} which uses the belief graph for effective planning and generalization. As used throughout this paper, the belief graph represents the state of the current game based on the textual description from the environment. Similar to commonsense knowledge, the belief graph can be fed to the agent as a full-graph (\textit{GATA\_Full}) or an evolve-graph (\textit{GATA\_Evolve}) and then aggregated as the current graph. It is worth noting that the full belief graph is considered as the ground truth state information in the TextWorld environment: it is the graph used by the TextWorld environment internally to modify the state information and the list of admissible actions. On the other hand, the evolve-belief graph is generated based on the observed state information. 
% given by the TextWorld environment as a textual description as suggested in GATA.

\begin{figure*}[t]
    \centering
    \includegraphics[scale=0.48]{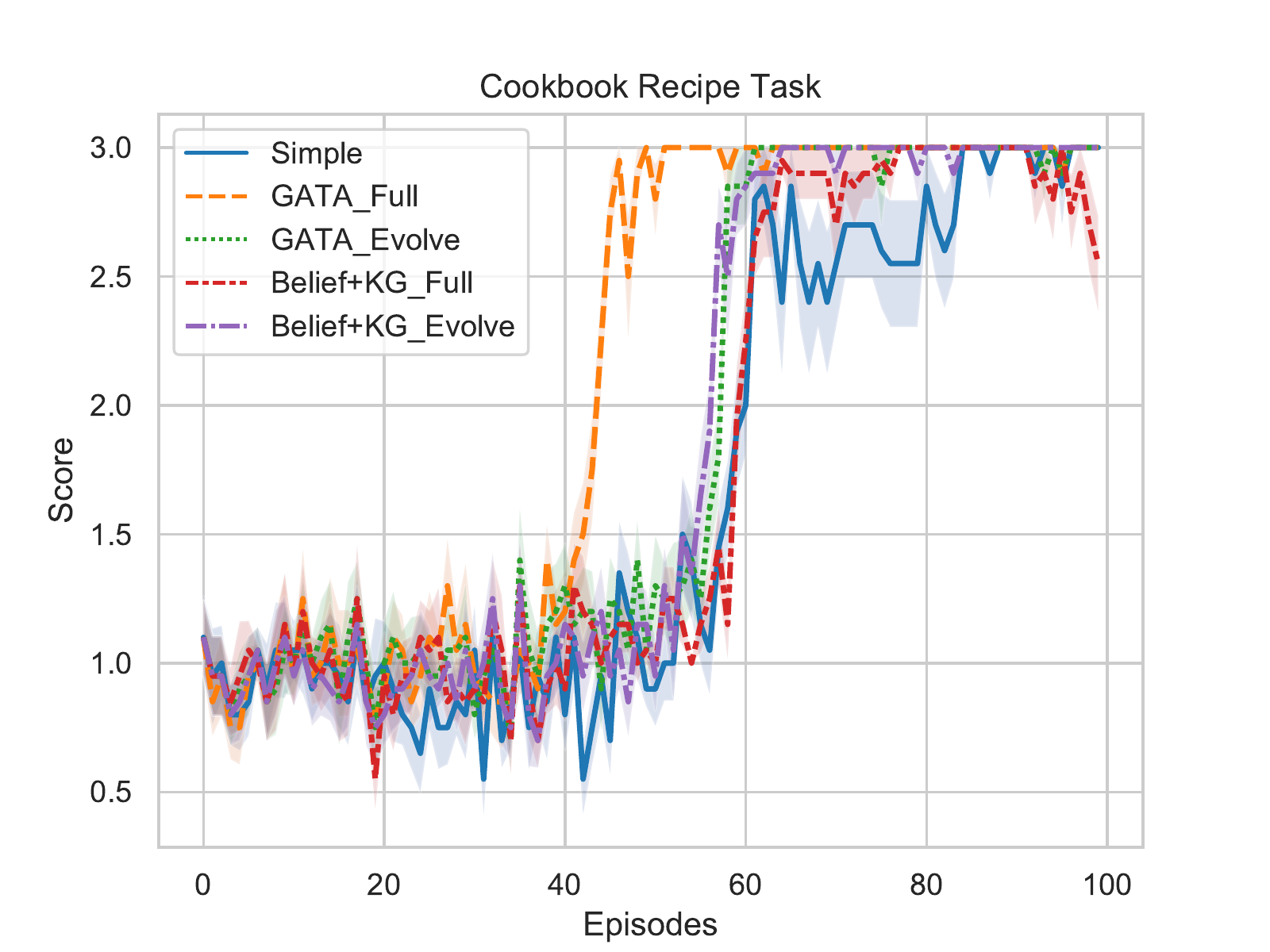}
    \includegraphics[scale=0.48]{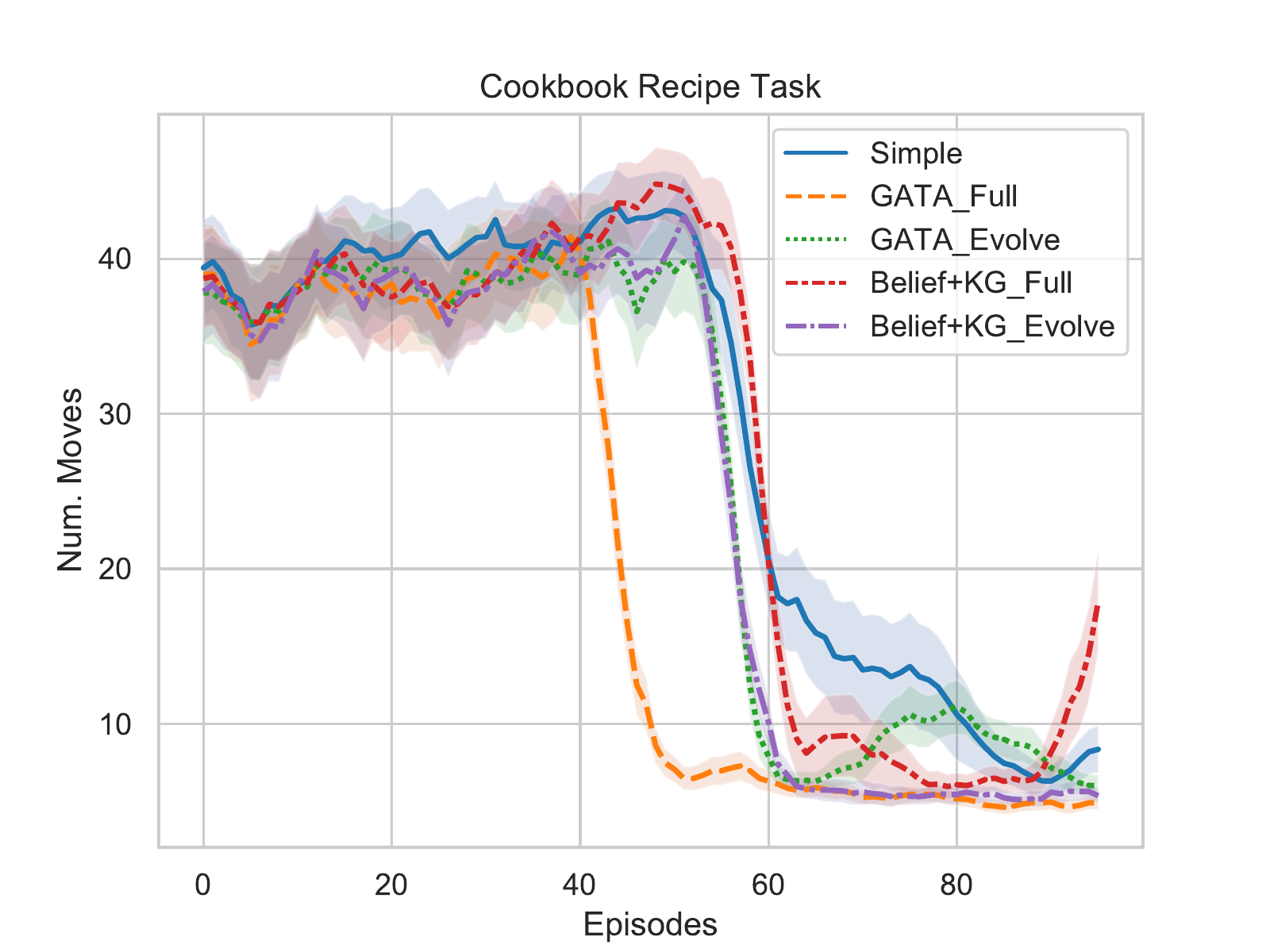}
    \caption{Comparison of agents for the Cooking Recipe task with belief graph and/or commonsense knowledge graph (averaged over $5$ runs).}
    \label{fig:cook_scores}
\end{figure*}

\subsubsection{Results on Cooking Recipe}

We compare both the simple and GATA agents against our agent which uses the commonsense knowledge extracted from ConceptNet. As before, we consider a full-graph setting and evolve-graph setting where either the full commonsense knowledge graph is available at the beginning of the game, or it is fed incrementally as the game proceeds, respectively.  For this task, we aggregate the commonsense knowledge graph with the belief graph (\textit{Belief+KG\_Full} and \textit{Belief+KG\_Evolve}). Figure~\ref{fig:cook_scores} shows the results for the Cooking recipe task averaged over $5$ runs and $20$ games, with $100$ episodes per run. As before, all the agents outperform the simple agent, which shows that using different state representations such as the belief graph and additional information such as commonsense knowledge improves the performance of an agent. 

\subsubsection{Discussion of Cooking Recipe}

We observe that the evolve-graph setting for both the GATA and Belief+KG performs better than the Belief+KG\_Full, as feeding more information can lead to noisy exploration as observed in the earlier task. 
More interestingly, we observe that GATA\_Full performs significantly better than the other agents. We believe that the reason for this result is the difficulty of the task at hand, and the process via which these cooking games are generated. Since the cooking recipe task (difficulty level 1) entails retrieving a single ingredient from the same room that the agent is present in, there are no meaningful concepts related to the current state that can be leveraged from the commonsense knowledge for better exploration. Even with the {\em difficult} task setting in this game environment (difficulty level $10$ with $3$ ingredients spread across $6$ rooms), the ingredients are randomly chosen and spread across the rooms. In such a game setting, the ground truth full belief graph is more beneficial than the commonsense knowledge graph. This is an interesting negative result, in that it shows that there can still be scenarios and domains where commonsense knowledge may not necessarily help an agent. We are actively exploring further settings of the cooking recipe task in order to understand and frame this effect better.

\section{Conclusion}

Previous approaches for text-based games like TextWorld primarily focused on text understanding and reinforcement learning for learning control policies and were thus sample inefficient. In contrast, humans utilize their commonsense knowledge to efficiently act in the world. As a step towards bridging this gap, we investigated the novel problem of using commonsense knowledge to build efficient RL agents for text-based games. We proposed a technique that symbolically represents the agent's belief of the world, and then combines that belief with commonsense knowledge from the ConceptNet knowledge graph in order to act in the world. We evaluated our approach on multiple tasks and environments and showed that commonsense knowledge can help the agent act efficiently and accurately. We also showcased some interesting negative results with respect to agents being overwhelmed with too much commonsense knowledge. We are currently actively studying this problem, and future work will report in more detail on this phenomenon.

\section*{Acknowledgments}

We thank Sadhana Kumaravel, Gerald Tesauro, and Murray Campbell for their feedback and help with this work.

{
\bibliographystyle{acl_natbib}
\bibliography{iqa}
}

\end{document}